\documentclass{article}
\usepackage{spconf,amsmath,graphicx}

\usepackage{amssymb}
\usepackage{booktabs}
\usepackage{tabularx}
\usepackage{multirow}

\usepackage{listings}
\usepackage{bbm}
\usepackage{algorithm}
\usepackage{comment}

\usepackage[normalem]{ulem}
\usepackage{url}

\title{\Ours{}: Decoupling and Dynamic Queries for Oriented Object Detection with Transformers}
\name{Qiang Zhou, Chaohui Yu, Zhibin Wang, Fan Wang
}
\address{Alibaba Group, Damo Academy, China}
\def\Ours{D$^2$Q-DETR}

\begin{document}
%

\maketitle
\begin{abstract}
   Despite the promising results, existing oriented object detection methods usually involve heuristically designed rules, e.g., RRoI generation, rotated NMS.
   In this paper, we propose an end-to-end framework for oriented object detection, which simplifies the model pipeline and obtains superior performance.
   Our framework is based on DETR, with the box regression head replaced with a points prediction head.
   The learning of points is more flexible, and the distribution of points can reflect the angle and size of the target rotated box.
   %
   We further propose to decouple the query features into classification and regression features, which significantly improves the model precision.
   Aerial images usually contain thousands of instances.
   To better balance model precision and efficiency, we propose a novel dynamic query design, which reduces the number of object queries in stacked decoder layers without sacrificing model performance.
   Finally, we rethink the label assignment strategy of existing DETR-like detectors and propose an effective label re-assignment strategy for improved performance.
   We name our method \Ours{}. Experiments on the largest and challenging DOTA-v1.0 and DOTA-v1.5 datasets
   show that \Ours{} outperforms existing NMS-based and NMS-free oriented object detection methods and achieves the new state-of-the-art.
\end{abstract}
\begin{keywords}
Rotated Object Detection, Transformer
\end{keywords}

\begin{figure*}[!h]
    \centering
    \includegraphics[width=0.9\textwidth]{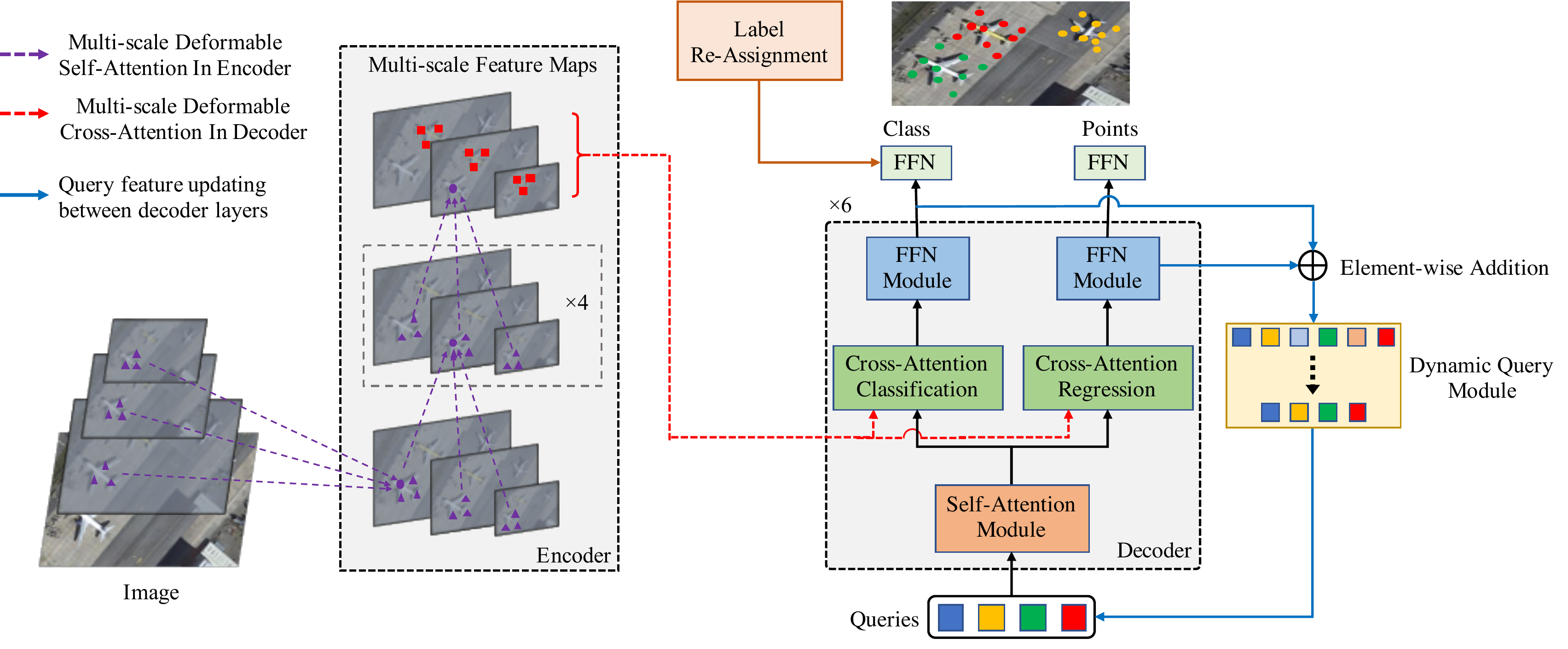}
    \vspace{-.1in}
    \caption{Overall architecture of the proposed \Ours{} for oriented object detection which consists of four main parts: the points prediction head, query feature decoupling, dynamic queries, and label re-assignment.}
    \label{fig:model}
\vspace{-.1in}
\end{figure*}

\section{Introduction}
\label{sec:intro}

Oriented object detection has recently attracted increasing attention
for its
important application
across different scenarios, including aerial images, scene text, and faces, etc.
Especially in aerial images, many well-designed oriented object detectors~\cite{roi_transformer,ConvexFA,dense_label_encoding,SCRDet,scrdet++,R3Det,ReDet,orientedrcnn}
have been proposed and reported promising results on large datasets~\cite{dota_dataset, dotav2_dataset}.
Compared with natural images, objects in aerial images usually 
exhibit 
dense distribution, large aspect ratios, and arbitrary orientations.
These characteristics make the model pipeline of the existing oriented object detectors complicated.
Our 
work here focuses on simplifying oriented object detection, eliminating the need for complex hand-crafted components, including but not limited to rule-based training target assignment, rotated RoI generation~\cite{roi_transformer,orientedrcnn}, rotated non-maximum suppression (NMS) and rotated RoI feature extractor~\cite{roi_transformer,ReDet}.

Our method is based on DETR~\cite{DETR}, which is a recently proposed end-to-end object detection framework using transformers.
The most relevant work to us is O$^2$DETR~\cite{o2detr}, which adds additional angle regression to DETR for oriented object detection.
As pointed out in~\cite{dense_label_encoding}, the direct regression of angle will cause two problems: one is boundary discontinuity caused by the periodicity of angle, and the other is the unit mismatch between box angle (in radians) and size (in pixels).
Unlike O$^2$DETR, 
we propose predicting representative points for each rotated box.
The learning of points is more flexible, and the distribution of points can reflect the angle and size of the target rotated box.
To our knowledge, 
we are the first to combine representative points prediction with transformers and achieve superior performance for oriented object detection.

It is common for aerial images to contain thousands of instances, which is different from natural images.
For example, in the DOTA-v1.0 dataset~\cite{dota_dataset}, the number of object instances per image can reach 2000. 
In order to improve model precision, we need to increase the number of queries, which will inevitably increase the computational complexity and memory usage of the model.
To better balance model precision and efficiency, in this work, we propose a novel \textbf{dynamic query} design, 
which gradually reduces the number of object queries in the stacked decoder layers without sacrificing model performance.

Many object detection works 
prior to 
DETR have shown that the decoupling of classification and regression features is beneficial in improving performance.
For example, Double Head~\cite{DoubleHead} proposes a double-head structure.
%
However, current DETR-like detectors do not consider the feature decoupling of classification and regression tasks, especially in the design of the decoder layer. 
In this work, we propose to decouple the query features into classification and regression features at the decoder layer.
To our knowledge, we are the first to consider \textbf{query feature decoupling} at the decoder layer for DETR-like detectors.

Furthermore, we rethink the label assignment of existing DETR-like detectors.
DETR-like detectors use bipartite matching between object queries and target boxes for label assignment and do not further impose any restrictions on the queries that match the target boxes. 
In contrast, 
%
we propose a \textbf{label re-assignment} strategy, 
which is used after bipartite matching to filter out low quality queries and effectively improves the model performance. 

All the above modules (i.e., points prediction head, query feature decoupling, dynamic queries, and label re-assignment) together constitute our end-to-end framework for oriented object detection,  
named \Ours{}. 
We achieve the new state-of-the-art on 
    challenging DOTA-v1.0 and DOTA-v1.5 datasets
    compared with existing NMS-based and NMS-free oriented object detection methods.

\section{Proposed Framework}
\label{sec:Method}

In this section, we illustrate our proposed \Ours{} for oriented object detection.

\subsection{Points Prediction Head}
%
As we stated in the introduction, 
to avoid problems raised by angle regression, our \Ours{} adopts an angle-free approach. 
Specifically, we predict a set of representative points  $\hat{\mathcal{C}}_i = \{ (\hat{x}_i^k, \hat{y}_i^k), ~ k=1, \cdots, K\}$ for each query, as shown in Fig.~\ref{fig:model}. 
$K$ denotes the number of predicted points for each query and is set to 9 by default.
The learning of points is more flexible, and the distribution of points can reflect the angle and size of the target box.
During inference, for each query, we convert the predicted point set $\hat{\mathcal{C}}_i$ to a rotated box with \textit{minAreaRect} function provided in OpenCV~\cite{opencv_library}.

\subsection{Query Feature Decoupling}

This section illustrates the proposed query feature decoupling, where the query features in the decoder layer are decoupled into two sets of class queries and box queries after the self-attention module, as shown in Fig.~\ref{fig:model}.
Class queries and box queries are responsible for object class prediction and points regression, respectively.
When stacking multiple decoder layers, the input query feature of each decoder layer is obtained by fusing the class-query feature and box-query feature of the previous decoder layer with element-wise addition.
In this way, the self-attention module also plays a role in exchanging information between class-query and box-query features.

\subsection{Dynamic Queries}
\label{sec:efficient_query}

%
DETR-like detectors stack multiple decoder layers, and each decoder layer uses the same number of object queries $N$.
The increase of $N$ will inevitably increase computational complexity and memory usage.
To better balance model performance and efficiency, we propose to dynamically reduce the number of queries in decoder layers.
The core idea is that we set the number of object queries of the first decoder layer to an initial value of $N_{\text{first}}$, while dynamically reducing the number of queries in subsequent decoder layers:
\begin{equation}
    N_i = (N_{\text{first}} - N_{\text{last}}) * \rho^{i} + N_{\text{last}},
    \label{eq:dynamic_query}
\end{equation}
where $N_{\text{first}}$ denotes the number of object queries in the first decoder layer, $N_{\text{last}}$ ($< N_{\text{first}}$) denotes the number of object queries in the last decoder layer, $N_i$ denotes the number of object queries in the $i$-th decoder layer. $\rho$ ($< 1$) is a hyper-parameter.
%
The prediction of class probability directly implies whether the object query corresponds to an existing object or background.
Hence,
we select the top $N_i$ object queries with larger class probability predictions in the $(i-1)$-th decoder layer.

\subsection{Query Label Re-Assignment}

\begin{figure}[!h]
    \centering
    \includegraphics[width=0.4\textwidth]{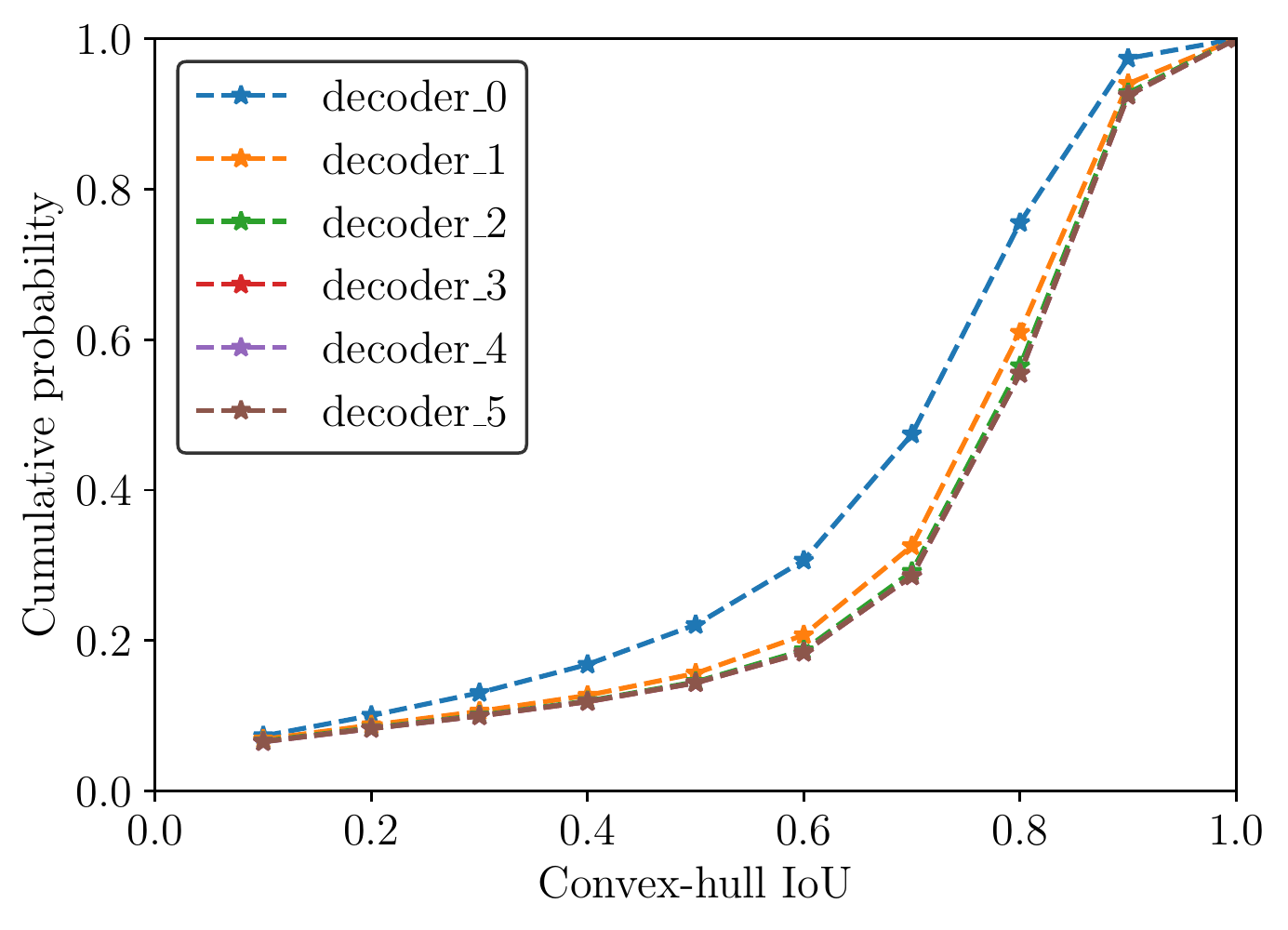}
    \vspace{-.1in}
    \caption{Cumulative distribution function plot of convex-hull IoU in different decoder layers.}
    \label{fig:cdf_iou}
\vspace{-.1in}
\end{figure}

To investigate the quality of queries after bipartite matching, we use a \Ours{} model after 50 epochs of training to statistic the convex-hull IoU between queries and matched target boxes on DOTA-v1.0 validation set.
As shown in Fig.~\ref{fig:cdf_iou}, in the last decoder layer, about 10\% queries and their matched target boxes have convex-hull IoU below 0.5.
And the ratio increases to about 25\% in the first decoder layer.
Obviously, the model performance and convergence speed will be affected by low-quality queries.
To alleviate the negative effect of low-quality queries, we propose query label re-assignment after bipartite matching.
As shown in Eq.~\ref{eq:reassign}, after bipartite matching, we check the convex-hull IoU between query predictions $\hat{\mathcal{C}}_i$ and matched target boxes $\mathcal{B}_{\hat{\sigma}(i)}$, and set the assigned label $y_{\hat{\sigma}(i)}$ of queries whose convex-hull IoU are lower than threshold $\tau$ to $\emptyset$ (\textit{no object} or background).

\begin{equation}
    y_{\hat{\sigma}(i)}^* = \begin{cases}
  y_{\hat{\sigma}(i)}, & \text{ if } ~
  {\frac{ | \Gamma(\hat{\mathcal{C}}_i) \cap \Gamma(\mathcal{B}_{\hat{\sigma}(i)}) | }{ | \Gamma(\hat{\mathcal{C}}_i) \cup \Gamma(\mathcal{B}_{\hat{\sigma}(i)}) | }} > \tau, \\
  \emptyset, & \text{ otherwise. }
\end{cases}
\label{eq:reassign}
\end{equation}
Where the hyper-parameter $\tau$ is set to 0.5 if not specified.
As stated in Eq.~\ref{eq:totalloss_reassign}, our label re-assignment only affects the classification loss $\mathcal{L}_{\text{cls}}$ with re-assigned class labels $y_{\hat{\sigma}(i)}^*$ .

\subsection{Training Loss}
In this section, we describe the training loss used in \Ours{}.
As shown in Eq.~\ref{eq:totalloss_reassign}, the total loss $\mathcal{L}$ includes one classification loss $\mathcal{L}_{\text{cls}}$ and two regression losses $\mathcal{L}_{L1}$ and $\mathcal{L}_{\text{iou}}$:
\begin{equation}
\label{eq:totalloss_reassign}
    \begin{split}
        \mathcal{L} = \frac{1}{N_{\text{pos}}} &\sum_{i=1}^{N} \Big[~ \lambda_{\text{cls}}~ \mathcal{L}_{\text{cls}} \big( \hat{\mathbf{p}}_i, y_{\hat{\sigma}(i)}^* \big) \\
        &+ \mathbbm{1}_{y_{\hat{\sigma}(i)} \neq \emptyset }~ \lambda_{L1}~ \mathcal{L}_{L1} \big(\Lambda(\hat{\mathcal{C}}_i), \Lambda(\mathcal{B}_{\hat{\sigma}(i)}) \big) \\
        &+ \mathbbm{1}_{y_{\hat{\sigma}(i)} \neq \emptyset }~ \lambda_{\text{iou}}~ \mathcal{L}_{\text{iou}} \big( \Gamma(\hat{\mathcal{C}}_i), \Gamma(\mathcal{B}_{\hat{\sigma}(i)}) \big) ~\Big].
    \end{split}
\end{equation}
where $\hat{\sigma}$ is the optimal one-to-one matching between the queries and the ground-truths without duplicates.
$N$ denotes the total number of queries, $N_{\text{pos}}$ denotes the number of queries matched with ground-truths.

\begin{figure}
    \small
    \centering
    \includegraphics[width=0.4\textwidth]{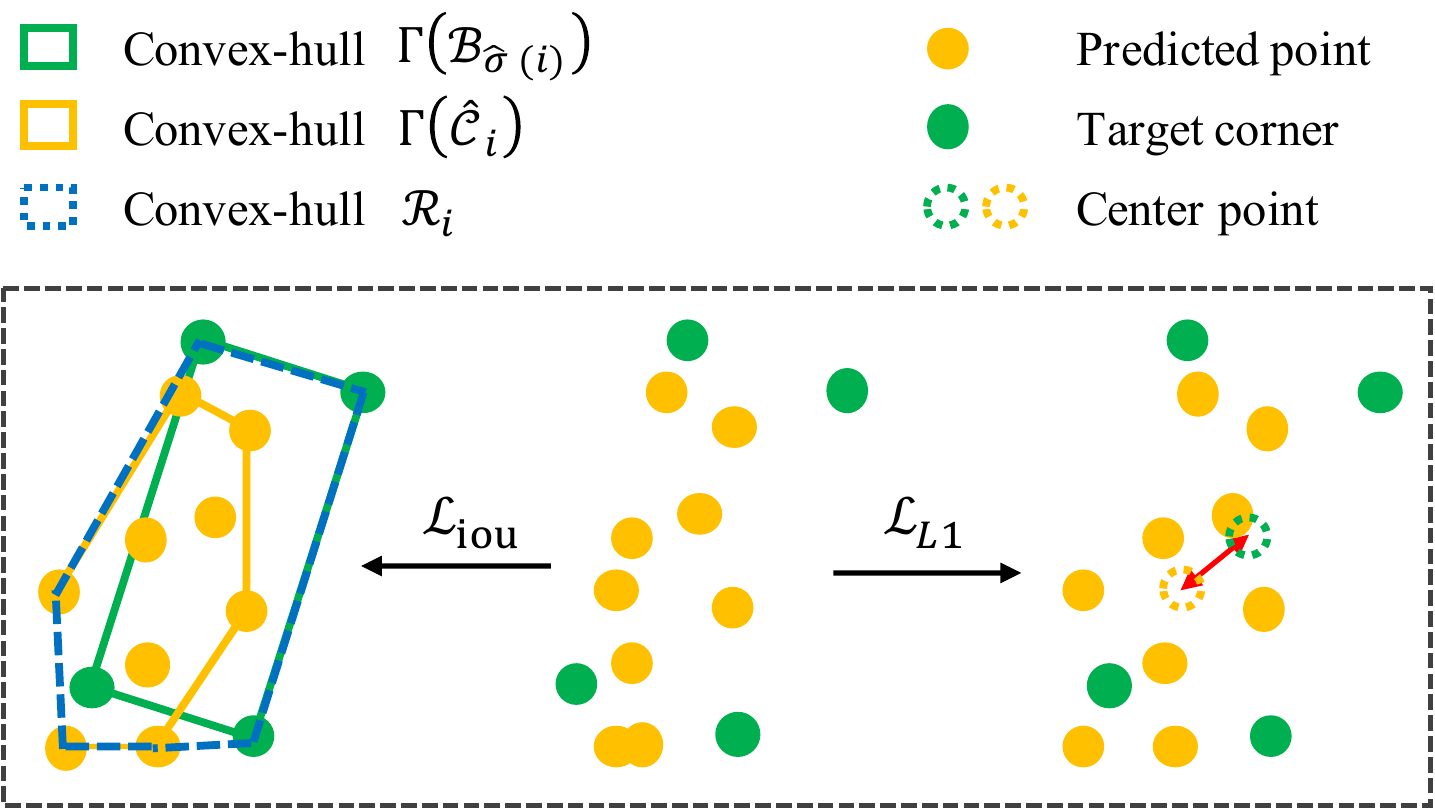}
    \vspace{-.1in}
    \caption{Schematic diagram of regression losses (including $\mathcal{L}_{L1}$ and $\mathcal{L}_{\text{iou}}$) for predicted points.}
    \label{fig:regression_loss}
    \vspace{-.1in}
\end{figure}

When calculating the classification loss $\mathcal{L}_{\text{cls}}$, we use the Focal Loss~\cite{focalloss}.
%
When calculating the regression losses (including $\mathcal{L}_{L1}$ and $\mathcal{L}_{\text{iou}}$), we make some changes since now we predict a point set for each query.
Concretely, we first use $\mathcal{L}_{L1}$ to narrow the distance between the two point sets of $\hat{\mathcal{C}}_i$ and $\mathcal{B}_{\hat{\sigma}(i)}$.
 As shown in Fig.~\ref{fig:regression_loss}, 
$\mathcal{L}_{L1}$ represents the $L$1 loss between the center points of the predicted point set $\hat{\mathcal{C}}_i$ and the target point set $\mathcal{B}_{\hat{\sigma}(i)}$. 
%
The defined $\mathcal{L}_{L1}$ loss helps the convergence of the learned point set, however, it can not drive the predicted point set $\hat{\mathcal{C}}_i$ to align well with the target bounding polygon $\mathcal{B}_{\hat{\sigma}(i)}$.
For better alignment of point sets, we further use the Convex-hull GIoU loss~\cite{ConvexFA} to compute $\mathcal{L}_\text{iou}$, which takes as input two convex hulls and can be formulated as:
\begin{equation}
    \begin{split}
        \mathcal{L}_{\text{iou}}\big( \Gamma(\hat{\mathcal{C}}_i), \Gamma(\mathcal{B}_{\hat{\sigma}(i)}) \big) &= 1 - \frac{\big| \Gamma(\hat{\mathcal{C}}_i) \cap \Gamma(\mathcal{B}_{\hat{\sigma}(i)}) \big|}{\big| \Gamma(\hat{\mathcal{C}}_i) \cup \Gamma(\mathcal{B}_{\hat{\sigma}(i)}) \big|} \\
        &+ \frac{\big|\mathcal{R}_i \setminus \big(\Gamma(\hat{\mathcal{C}}_i) \cup \Gamma(\mathcal{B}_{\hat{\sigma}(i)}) \big) \big|}{\big| \mathcal{R}_i \big|},
    \end{split}
    \label{eq:loss_iou}
\end{equation}
where $\Gamma(\cdot)$ represents the Jarvis March algorithm~\cite{convexhull} to convert a point set to a minimal convex-hull.
$\mathcal{R}_i$ is the minimum bounding polygon of convex-hulls $\Gamma(\hat{\mathcal{C}}_i)$ and $\Gamma(\mathcal{B}_{\hat{\sigma}(i)})$, as shown in Fig.~\ref{fig:regression_loss}.

\begin{figure*}[!h]
    \centering
    \small
    \includegraphics[width=0.98\textwidth]{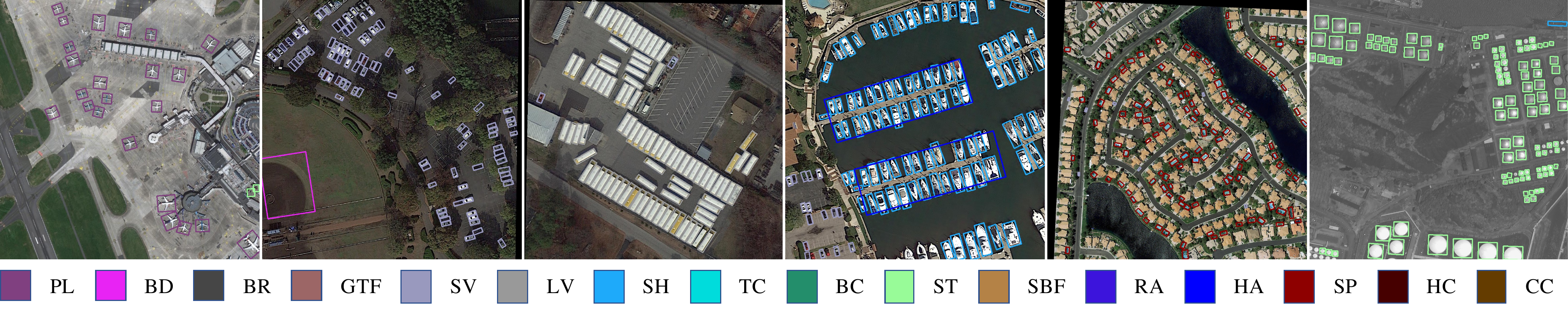}
    \vspace{-.1in}
    \caption{Examples of detection results on the DOTA-v1.5 test set.
    }
    \label{fig:vis_result}
    \vspace{-.1in}
\end{figure*}

\section{Experiments}

\subsection{Dataset}

DOTA~\cite{dota_dataset} is the largest dataset for oriented object detection
in aerial images with two commonly used versions: \textbf{DOTA-v1.0} and \textbf{DOTA-v1.5}. 
\textbf{DOTA-v1.0} contains 2806 large
aerial images with the size ranges from $800 \times 800$ to
$4000 \times 4000$ and 188, 282 instances among 15 common categories.
\textbf{DOTA-v1.5} is released for DOAI Challenge
2019 with a new category and more
extremely small instances (less than 10 pixels). DOTA-v1.5
contains 402, 089 instances. Compared with DOTA-v1.0,
DOTA-v1.5 is more challenging but stable during training.
%
%


\subsection{Implementation Details}
We use 8 V100 GPUs with a total batch size of 8 for training.
AdamW~\cite{adam} optimizer is adopted with the base learning rate of $1.5\times 10^{-4}$, $\beta_1=0.9$, $\beta_2=0.999$, and weight decay of 0.05.
Like DETR, we use six encoder layers, six decoder layers, and multi-scale features with downsampling rates of 8, 16, 32, and 64.
In the inference phase, we use OpenCV's \textit{minAreaRect} function to convert the predicted points of each object query into a rotated box. No other post-processing operations, e.g., rotated NMS, are required.

\subsection{Comparison with the State-of-the-art}

\vspace{-.1in}
\begin{table}[!h]
\centering 
    \resizebox{0.35\textwidth}{!}{
    \begin{tabular}{r|c|c}
    \toprule
    Method                   & Backbone & mAP (\%)  \\ \midrule
    \textbf{NMS-based method}    &   &  \\
    RoI Trans.$^{*}$~\cite{roi_transformer} & R101-FPN & 69.56 \\ 
    O$^2$-DNet$^{*}$~\cite{o2dnet}  & H104 & 72.80 \\
    DRN$^{*}$~\cite{DRN} & H104 & 73.23 \\
    Gliding Vertex$^{*}$~\cite{GlidingVertex} & R101-FPN & 75.02 \\ 
    R$^3$DET$^*$~\cite{R3Det} & R-152 & 76.47 \\
    SCRDet++$^{*}$~\cite{scrdet++}  & R152-FPN & 76.56 \\
    CFA$^*$~\cite{ConvexFA} & R-152 & 76.67 \\
    R$^3$DET-DCL$^*$~\cite{dense_label_encoding} & R-152 & 77.37 \\
    HSP$^*$~\cite{HSP2020} & R-101 & 78.01 \\
    ReDet$^{*}$~\cite{ReDet} & ReR50-ReFPN & 80.10 \\ \midrule
    \textbf{NMS-free method}   &   & \\
    O$^2$DETR$^{*\sharp}$~\cite{o2detr}   & R50-FPN & 72.15 \\
    
    \Ours{}$^*$ (Ours)         & R50-FPN  & 76.58  \\
    \Ours{}$^*$ (Ours)         & Swin-L-FPN &79.60 \\
    \Ours{}$^{\dag*}$ (Ours) & R50-FPN & 79.40 \\
    \Ours{}$^{\dag*}$ (Ours) & Swin-L-FPN & \textbf{81.24} \\
    \bottomrule
    \end{tabular}
}
\caption{Performance comparisons on DOTA-v1.0 test set. 
$^*$ denotes multi-scale training and testing. 
$^\dag$ indicates that we finetune an additional RCNN head to compare with two-stage methods like ReDet~\cite{ReDet}.
}
\label{tbl:dota_1.0}
\vspace{-.1in}
\end{table}

\begin{table}[!h]
\centering 
    \resizebox{0.35\textwidth}{!}{
    \begin{tabular}{r|c|c}
    \toprule
    Method   & Backbone & mAP (\%)  \\ \midrule
    \textbf{NMS-based method}    &   & \\
    
    RetinaNet-O~\cite{focalloss} & R50-FPN  & 59.16 \\
    FR-O~\cite{dota_dataset} & R50-FPN  & 62.00 \\
    Mask R-CNN~\cite{maskrcnn} & R50-FPN  &  62.67 \\
    HTC~\cite{HTC} & R50-FPN  &  63.40 \\
    OWSR$^{*}$~\cite{owsr2019} & R101-FPN  & 74.90 \\
    ReDet$^{*}$~\cite{ReDet} & ReR50-ReFPN  & 76.80 \\
    \midrule
    
    \textbf{NMS-free method}    &  & \\
    \Ours{}$^*$ (Ours) & Swin-L-FPN  & 75.23 \\
    \Ours{}$^{\dag*}$ (Ours) & Swin-L-FPN & \textbf{78.85} \\
    \bottomrule
    \end{tabular}
}
\caption{Performance comparisons on DOTA-v1.5 test set. 
}
\label{tbl:dota_1.5}
\vspace{-.1in}
\end{table}


\textbf{Results on DOTA-v1.0.}
As shown in Table~\ref{tbl:dota_1.0}, we compare our \Ours{} with existing methods on DOTA-v1.0 OBB Task
and achieve the new state-of-the-art performance of \textbf{81.24} mAP on the test set.

\textbf{Results on DOTA-v1.5.}
Compared with DOTA-v1.0, DOTA-v1.5 contains many extremely small instances, which increases the difficulty of object detection. 
We report OBB results on DOTA-v1.5 test set in Table~\ref{tbl:dota_1.5}.
Compared to the previous best result of ReDet~\cite{ReDet}, our
model achieves state-of-the-art performance of \textbf{78.85} mAP.
Fig.~\ref{fig:vis_result} shows some detection results on the DOTA-v1.5 dataset.

\subsection{Ablations}
\begin{table}[!h]
\centering
    \resizebox{0.48\textwidth}{!}{
    \begin{tabular}{@{}c|c|c|c|l|l@{}}
    \toprule
    Points Head  & Decoupling-Q    & LR  & Dynamic-Q & mAP (\%) & GFLOPs  \\ \midrule
                 &              &    &    & 70.72 & 194.52            \\ 
    $\checkmark$ &              &    &    & 71.44 ($\uparrow$)  & 194.60 \\ 
    $\checkmark$ & $\checkmark$ &    &    & 73.25 ($\uparrow$) & 211.44  \\ 
    $\checkmark$ & $\checkmark$ &  $\checkmark$  & &  74.66 ($\uparrow$)  & 211.44  \\ 
    $\checkmark$ & $\checkmark$ &  $\checkmark$  &  $\checkmark$  & 74.57 & 197.63 ($\downarrow$) \\ 
    \bottomrule
    \end{tabular}
    }
    \caption{
    Effectiveness of proposed components including points prediction for oriented box (\textbf{Points Head}), decoupling query features (\textbf{Decoupling-Q}), dynamic query design (\textbf{Dynamic-Q}) and query label re-assignment (\textbf{LR}).
    }
    \label{tbl:factor_by_factor}
    \vspace{-.1in}
\end{table}

We conduct ablation experiments on the DOTA-v1.0 dataset, with the training set for training and the validation set for testing.
All models are trained for 50 epochs and the learning rate is decayed at the 40-th epoch by a factor of 0.1. 
%
%
As shown in Table~\ref{tbl:factor_by_factor}, each component we proposed positively impacts the model performance, and combining all these components achieves the best performance.

\section{Conclusion}

We have
proposed an end-to-end detector, named \Ours{}, for arbitrary-oriented object detection.
%
%
%
%
Experimental results on DOTA-v1.0 and DOTA-v1.5 datasets demonstrate the superiority of our \Ours{} framework against current advanced NMS-based and NMS-free detectors. 

\bibliographystyle{IEEEbib}
\bibliography{strings,refs}

\begin{thebibliography}{10}

\bibitem{roi_transformer}
Jian Ding, Nan Xue, Yang Long, Gui{-}Song Xia, and Qikai Lu,
\newblock ``Learning roi transformer for oriented object detection in aerial
  images,''
\newblock in {\em CVPR}, 2019.

\bibitem{ConvexFA}
Zonghao Guo, Chang Liu, Xiaosong Zhang, Jianbin Jiao, Xiangyang Ji, and Qixiang
  Ye,
\newblock ``Beyond bounding-box: Convex-hull feature adaptation for oriented
  and densely packed object detection,''
\newblock in {\em CVPR}, 2021.

\bibitem{dense_label_encoding}
Xue Yang, Liping Hou, Yue Zhou, Wentao Wang, and Junchi Yan,
\newblock ``Dense label encoding for boundary discontinuity free rotation
  detection,''
\newblock in {\em CVPR}, 2021.

\bibitem{SCRDet}
Xue Yang, Jirui Yang, Junchi Yan, Yue Zhang, Tengfei Zhang, Zhi Guo, Xian Sun,
  and Kun Fu,
\newblock ``Scrdet: Towards more robust detection for small, cluttered and
  rotated objects,''
\newblock in {\em ICCV}, 2019.

\bibitem{scrdet++}
Xue Yang, Junchi Yan, Xiaokang Yang, Jin Tang, Wenlong Liao, and Tao He,
\newblock ``Scrdet++: Detecting small, cluttered and rotated objects via
  instance-level feature denoising and rotation loss smoothing,''
\newblock {\em CoRR}, vol. abs/2004.13316, 2020.

\bibitem{R3Det}
Xue Yang, Junchi Yan, Ziming Feng, and Tao He,
\newblock ``R3det: Refined single-stage detector with feature refinement for
  rotating object,''
\newblock in {\em AAAI}, 2021.

\bibitem{ReDet}
Jiaming Han, Jian Ding, Nan Xue, and Gui{-}Song Xia,
\newblock ``Redet: {A} rotation-equivariant detector for aerial object
  detection,''
\newblock in {\em CVPR}, 2021.

\bibitem{orientedrcnn}
Xingxing Xie, Gong Cheng, Jiabao Wang, Xiwen Yao, and Junwei Han,
\newblock ``Oriented {R-CNN} for object detection,''
\newblock in {\em ICCV}, 2021.

\bibitem{dota_dataset}
Gui{-}Song Xia, Xiang Bai, Jian Ding, Zhen Zhu, Serge~J. Belongie, Jiebo Luo,
  Mihai Datcu, Marcello Pelillo, and Liangpei Zhang,
\newblock ``{DOTA:} {A} large-scale dataset for object detection in aerial
  images,''
\newblock in {\em CVPR}, 2018.

\bibitem{dotav2_dataset}
Jian Ding, Nan Xue, Gui{-}Song Xia, Xiang Bai, Wen Yang, Michael~Ying Yang,
  Serge~J. Belongie, Jiebo Luo, Mihai Datcu, Marcello Pelillo, and Liangpei
  Zhang,
\newblock ``Object detection in aerial images: {A} large-scale benchmark and
  challenges,''
\newblock {\em CoRR}, vol. abs/2102.12219, 2021.

\bibitem{DETR}
Nicolas Carion, Francisco Massa, Gabriel Synnaeve, Nicolas Usunier, Alexander
  Kirillov, and Sergey Zagoruyko,
\newblock ``End-to-end object detection with transformers,''
\newblock in {\em ECCV}, Andrea Vedaldi, Horst Bischof, Thomas Brox, and
  Jan{-}Michael Frahm, Eds., 2020.

\bibitem{o2detr}
Teli Ma, Mingyuan Mao, Honghui Zheng, Peng Gao, Xiaodi Wang, Shumin Han, Errui
  Ding, Baochang Zhang, and David~S. Doermann,
\newblock ``Oriented object detection with transformer,''
\newblock {\em CoRR}, vol. abs/2106.03146, 2021.

\bibitem{DoubleHead}
Yue Wu, Yinpeng Chen, Lu~Yuan, Zicheng Liu, Lijuan Wang, Hongzhi Li, and Yun
  Fu,
\newblock ``Rethinking classification and localization for object detection,''
\newblock in {\em CVPR}. 2020, Computer Vision Foundation / {IEEE}.

\bibitem{opencv_library}
G.~Bradski,
\newblock ``{The OpenCV Library},''
\newblock {\em Dr. Dobb's Journal of Software Tools}, 2000.

\bibitem{focalloss}
Tsung{-}Yi Lin, Priya Goyal, Ross~B. Girshick, Kaiming He, and Piotr
  Doll{\'{a}}r,
\newblock ``Focal loss for dense object detection,''
\newblock in {\em ICCV}, 2017.

\bibitem{convexhull}
Ray~A. Jarvis,
\newblock ``On the identification of the convex hull of a finite set of points
  in the plane,''
\newblock {\em Inf. Process. Lett.}, vol. 2, no. 1, pp. 18--21, 1973.

\bibitem{adam}
Diederik~P. Kingma and Jimmy Ba,
\newblock ``Adam: {A} method for stochastic optimization,''
\newblock in {\em ICLR}, Yoshua Bengio and Yann LeCun, Eds., 2015.

\bibitem{o2dnet}
Haoran Wei, Yue Zhang, Zhonghan Chang, Hao Li, Hongqi Wang, and Xian Sun,
\newblock ``Oriented objects as pairs of middle lines,''
\newblock {\em ISPRS Journal of Photogrammetry and Remote Sensing}, vol. 169,
  pp. 268--279, 2020.

\bibitem{DRN}
Xingjia Pan, Yuqiang Ren, Kekai Sheng, Weiming Dong, Haolei Yuan, Xiaowei Guo,
  Chongyang Ma, and Changsheng Xu,
\newblock ``Dynamic refinement network for oriented and densely packed object
  detection,''
\newblock in {\em CVPR}, 2020.

\bibitem{GlidingVertex}
Yongchao Xu, Mingtao Fu, Qimeng Wang, Yukang Wang, Kai Chen, Gui{-}Song Xia,
  and Xiang Bai,
\newblock ``Gliding vertex on the horizontal bounding box for multi-oriented
  object detection,''
\newblock {\em IEEE TPAMI}, pp. 1452--1459, 2021.

\bibitem{HSP2020}
Chunyan Xu, Chengzheng Li, Zhen Cui, Tong Zhang, and Jian Yang,
\newblock ``Hierarchical semantic propagation for object detection in remote
  sensing imagery,''
\newblock {\em {IEEE} Trans. Geosci. Remote. Sens.}, vol. 58, no. 6, pp.
  4353--4364, 2020.

\bibitem{maskrcnn}
Kaiming He, Georgia Gkioxari, Piotr Doll{\'{a}}r, and Ross~B. Girshick,
\newblock ``Mask {R-CNN},''
\newblock in {\em ICCV}, 2017.

\bibitem{HTC}
Kai Chen, Jiangmiao Pang, Jiaqi Wang, Yu~Xiong, Xiaoxiao Li, Shuyang Sun,
  Wansen Feng, Ziwei Liu, Jianping Shi, Wanli Ouyang, Chen~Change Loy, and
  Dahua Lin,
\newblock ``Hybrid task cascade for instance segmentation,''
\newblock in {\em CVPR}, 2019.

\bibitem{owsr2019}
Chengzheng Li, Chunyan Xu, Zhen Cui, Dan Wang, Zequn Jie, Tong Zhang, and Jian
  Yang,
\newblock ``Learning object-wise semantic representation for detection in
  remote sensing imagery,''
\newblock in {\em CVPRW}, 2019.

\end{thebibliography}

\end{document}